\documentclass[letterpaper]{article} 
\usepackage{aaai2026}
\usepackage{times}  
\usepackage{helvet}  
\usepackage{courier}  
\usepackage[hyphens]{url}  
\usepackage{graphicx} 
\usepackage{amsmath}
\usepackage{amssymb}
\usepackage{booktabs}
\usepackage{natbib}  
\usepackage{caption} 
\usepackage{algorithm}
\usepackage{algorithmic}

\usepackage{newfloat}
\usepackage{listings}
\DeclareCaptionStyle{ruled}{labelfont=normalfont,labelsep=colon,strut=off} 
\floatstyle{ruled}
\newfloat{listing}{tb}{lst}{}
\floatname{listing}{Listing}
\newcommand\shortdots{\makebox[1em][c]{.\hfil.\hfil.}}

\usepackage[hidelinks]{hyperref}

\title{Object-centric Denoising Diffusion Models for Physical Reasoning}
\author {
    Moritz Lange\textsuperscript{\rm 1},
    Raphael C. Engelhardt\textsuperscript{\rm 2},
    Wolfgang Konen\textsuperscript{\rm 2},
    Andrew Melnik\textsuperscript{\rm 3},
    Laurenz Wiskott\textsuperscript{\rm 1}
}
\affiliations {
    \textsuperscript{\rm 1}Faculty of Computer Science, Ruhr University Bochum, Bochum, Germany\\
    \textsuperscript{\rm 2}TH Köln, Köln, Germany\\
    \textsuperscript{\rm 3}Bielefeld University, Bielefeld, Germany\\
    moritz.lange@rub.de, raphael.engelhardt@th-koeln.de, wolfgang.konen@th-koeln.de,\\ andrew.melnik.papers@gmail.com, laurenz.wiskott@rub.de
}

\usepackage{bibentry}

\begin{document}

\maketitle

\begin{abstract}
Reasoning about the trajectories of multiple, interacting objects is integral to physical reasoning tasks in machine learning. This involves conditions imposed on the objects at different time steps, for instance initial states or desired goal states. Existing approaches in physical reasoning generally rely on autoregressive modeling, which can only be conditioned on initial states, but not on later states. In fields such as planning for reinforcement learning, similar challenges are being addressed with denoising diffusion models. In this work, we propose an object-centric denoising diffusion model architecture for physical reasoning that is translation equivariant over time, permutation equivariant over objects, and can be conditioned on arbitrary time steps for arbitrary objects. We demonstrate how this model can solve tasks with multiple conditions and examine its performance when changing object numbers and trajectory lengths during inference.
\end{abstract}

\section{Introduction}

Physical reasoning, in machine learning, often requires the prediction of object trajectories. In many scenarios, it is useful to imagine trajectories based on multiple conditions: In reinforcement learning, an agent might want to move objects to achieve a certain goal. In a counterfactual question scenario, a model might seek to explain what would have happened if certain conditions had been different. Such scenarios usually involve multiple objects that interact with each other, for instance through collisions or friction. To foster development on these physical reasoning challenges, a number of interactive or counterfactual benchmarks have been developed \citep{melnik_benchmarks_2023}.

The conditions imposed on objects often affect various time steps in trajectories. Some objects might have a fixed initial state, while there are desired goal states for other objects. At the same time, the system retains certain degrees of freedom, which can be controlled by a model or agent.

\begin{figure}[ht]
\centering
\includegraphics[width=\columnwidth]{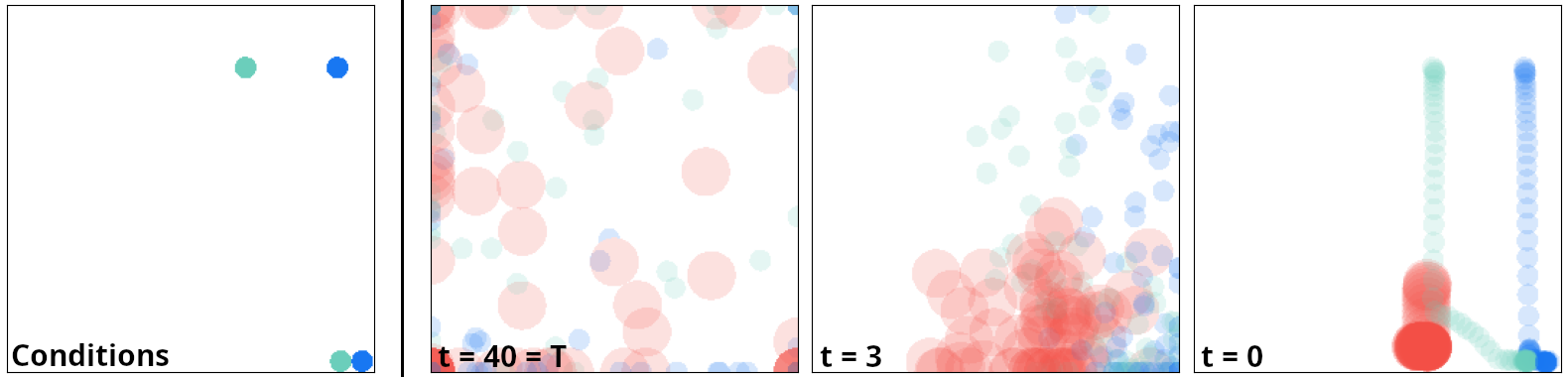}

\caption{\textit{Left:} Provided conditions fix initial and final states of green and blue ball, no conditions on the red ball. \textit{Right:} The denoising process. The model is able to place the red ball so that conditions are fulfilled. Object positions (centers) are clipped to image boundaries for display purposes.}
\label{fig:denoised_traj}
\end{figure}

Such an every-day physical setup, however, still presents unique challenges: There are many potential object interactions, and conditioning on various time steps prevents the use of regular autoregressive modeling approaches such as conventional differential equations or autoregressive generative models. Nevertheless, existing reasoning approaches \citep{girdhar_forward_2020,wu_slotformer_2023,daniel_ddlp_2023} often rely on autoregressive forward models. These are conditioned on initial states only and require a trial and error approach: Values for the degrees of freedom are sampled and the corresponding trajectories modeled until desired goal states are reached.

Building on denoising diffusion models for trajectory prediction of agents \citep{janner_planning_2022,ye_diffusion_2024}, we propose an object-centric denoising diffusion model for physical reasoning. It generates all time steps of the trajectories of multiple objects simultaneously, and can thus consider arbitrary conditions for different objects and time steps. Unlike existing denoising diffusion models -- which lack an object-centric design -- our model is translation equivariant over time and permutation equivariant over objects. This also grants it flexibility regarding trajectory lengths and object numbers during inference (within the limits of the training data distribution). We additionally employ two new methods to infuse and enforce conditions, so that our model can, in principle, solve physical reasoning scenarios with arbitrary conditions and degrees of freedom. Figure \ref{fig:denoised_traj} illustrates an example scenario. Overall, our \textbf{contributions} are:

\begin{itemize}
    \item We propose a novel object-centric denoising diffusion model architecture for physical reasoning, which models temporal evolution and interactions on an object-level.
    \item We propose and combine two novel ways of conditioning the denoising diffusion model: A soft conditioning approach that provides conditions to the model through masked interpolation, and a hard conditioning approach that shifts generated trajectories to enforce conditions.
\end{itemize}

\noindent Our implementation is publicly available on GitHub\footnote{\href{https://github.com/wiskott-lab/diffusion4phyre}{https://github.com/wiskott-lab/diffusion4phyre}}.

\section{Background}

The following gives an overview and brief introduction into relevant concepts of physical reasoning and denoising diffusion.

\subsection{Physical Reasoning}

Physical reasoning models make predictive, descriptive, counterfactual or explanatory statements, or perform planning, in a physical environment. State of the art in literature are modular setups that first process inputs (often visual), then perform the actual reasoning, and finally compute a solution to the present task \citep{melnik_benchmarks_2023}. The reasoning components are generally recurrent neural networks or transformers that model the temporal evolution of objects autoregressively. Recent models are object-centric, i.\,e.\ they decompose a scene into individual separate objects, and permutation equivariant with respect to the objects \citep{daniel_ddlp_2023, wu_slotformer_2023}. In other words, they treat each object individually -- an important inductive bias for physical reasoning. If object interactions are modeled explicitly, this is commonly done with graph neural networks \cite{li_deconfounding_2022} or with transformers \cite{qi_learning_2020}.

PHYRE \citep{bakhtin_phyre_2019} is a popular physical reasoning benchmark. It presents two dimensional vertical scenes of multiple distinct, rigid objects. These have one of three shapes (ball, bar, cup) and different sizes, and are subject to gravitation and friction. Each scenario is specified through the initial states of all objects, except one (PHYRE-B) or two (PHYRE-2B) red balls. A simulator models the scene's temporal evolution with objects falling and colliding. The task is to place the red balls, so that a green object will touch a blue or purple object for $>3$ seconds. Some objects (red, green, blue, grey) can freely move, while others (purple, black) are fixed. In PHYRE terminology, a pre-defined initial object configuration is called the \textit{task}. A \textit{template} contains multiple tasks with similar but slightly different configurations. Selecting location and size of the red ball (we only consider PHYRE-B here) is called an \textit{action}.

Solution approaches often rely on forward modeling of object trajectories \citep{qi_learning_2020,wu_slotformer_2023,daniel_ddlp_2023}. Their autoregressive nature restricts them to a trial and error approach for finding appropriate actions.

In this work, we present a general conditional denoising diffusion model for object dynamics. We use PHYRE benchmark scenarios to evaluate our model, and find it to be quite challenging -- especially, as it turns out, due to their limited coverage of possible interactions. To maintain a focus on dynamics modeling, we disregard some aspects of PHYRE: We use the available object feature vectors rather than image data. Additionally, we do not attempt the actual task of placing the red ball: While our model can find a suitable initial position for the red ball (see Figure \ref{fig:denoised_traj}), deploying it automatically would require an additional module to generate absolute (rather than relative) values for goal states.

\subsection{Denoising Diffusion Models}

\begin{figure*}[ht]
\centering
\includegraphics[width=\textwidth]{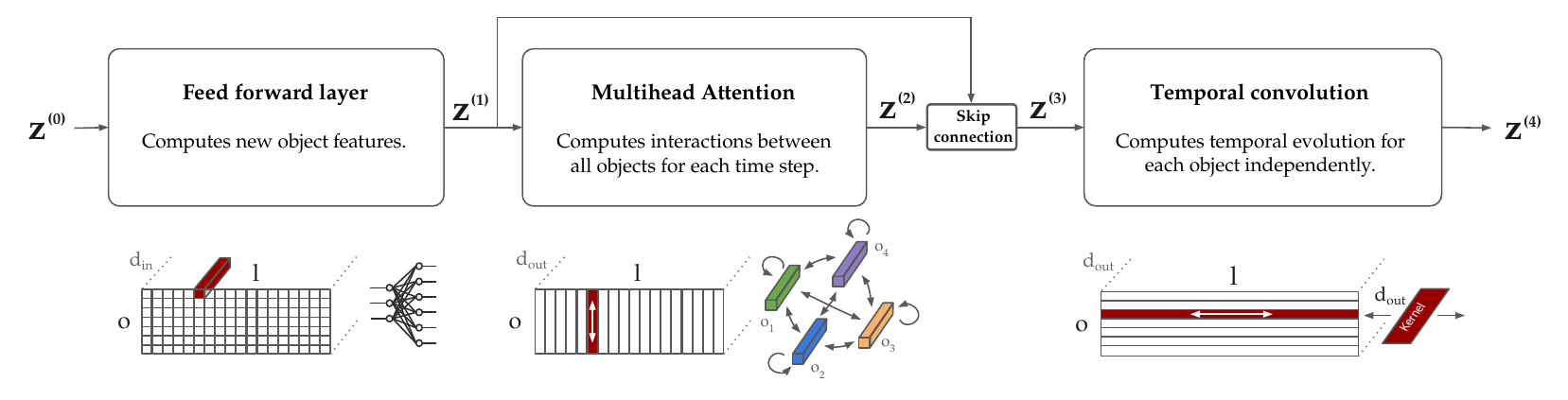}

\caption{Attention-convolution block. The variables are defined as follows: $o$ is the index of an object, $l$ is the timestep in the trajectory, $d$ is the number of object features.}
\label{fig:acb}
\end{figure*}

Denoising diffusion models are generative models that approximate data through a denoising process \citep{ho_denoising_2020, chan_tutorial_2024}
\[
p_\theta\left(\boldsymbol{x}^{[0]}\right) = \int p(\boldsymbol{x}^{[T]}) \prod_{t=1}^T p_\theta \! \left(\boldsymbol{x}^{[t-1]} \mid  \boldsymbol{x}^{[t]}\right) \mathrm{d}\boldsymbol{x}^{[1:T]}\, .
\]
This reverses a forward process of $t \in \{0, \shortdots, T\}$ steps, in which Gaussian noise is added repeatedly to original data $\boldsymbol{x}^{[0]}$ to obtain a noisy $\boldsymbol{x}^{[T]}$. In our model, $\boldsymbol{x}^{[T]}$ becomes a zero-mean, isotropic Gaussian.
The backward denoising process can then generate a new $\boldsymbol{x}^{[0]}$, similar to those in the training data, from random noise. It relies on a model, parameterized by $\theta$, to compute

\[
p_\theta\left(\boldsymbol{x}^{[t-1]}\mid \boldsymbol{x}^{[t]}\right) = \mathcal{N}\left(\boldsymbol{x}^{[t-1]}; \boldsymbol{\mu}_\theta\left(\boldsymbol{x}^{[t]}, \boldsymbol{x}^{[0]}\right), \tilde{\beta}^{[t]} \boldsymbol{I}\right) \, .
\]

Parameters $\tilde{\beta}^{[t]}$ relate to a noise schedule that increases from 0 to 1 with increasing $t$. For our model, we adopt the cosine schedule proposed by \citet{nichol_improved_2021}.

Denoising is an iterative process and $\boldsymbol{x}^{[t]}$ is given.
To compute $\boldsymbol{\mu}_\theta$, a neural network is usually employed to predict $\boldsymbol{x}^{[0]}$ from $\boldsymbol{x}^{[t]}$. Alternatively, the network can predict the Gaussian noise $\boldsymbol{\varepsilon}^{[t]}$ that the forward process would have added to $\boldsymbol{x}^{[0]}$ to obtain $\boldsymbol{x}^{[t]}$. In practice, we predict a superposition of both to improve stability \citep{salimans_progressive_2022}. Either quantity can be used to obtain $\boldsymbol{\mu}_\theta$. For simplicity, we consider $\boldsymbol{x}^{[0]}$ as the prediction target in the following.

The neural network architectures $\theta: \left(\boldsymbol{x}^{[t]}, t\right) \mapsto \boldsymbol{x}^{[0]}$ that are commonly employed in denoising diffusion models are U-Nets \citep{ronneberger_u-net_2015}, used e.\,g.\ in \citet{ho_denoising_2020, janner_planning_2022} and recently also transformers \citep{vaswani_attention_2017}, used e.\,g.\ in \citet{tashiro_csdi_2021, peebles_scalable_2023}.

\paragraph{Conditioning}
Targeted data generation requires conditioning the denoising diffusion model. Classifier-based guidance \citep{dhariwal_diffusion_2021} conditions through the gradients of a learned classifier, an approach suited more for image generation than trajectory generation. Classifier-free guidance removes the need of a classifier \citep{ho_classifier-free_2021}. The diffusion model itself is instead conditioned on an additional input $c$, which is filled with a placeholder $\o$ in the absence of a condition. Sampling uses the combination

\begin{align*}
    w \, \boldsymbol{x}^{[0]}_\theta \!\! \left(\boldsymbol{x}^{[t]}, t, c\right) + (1-w) \, \boldsymbol{x}^{[0]}_\theta \!\! \left(\boldsymbol{x}^{[t]}, t, \o\right) ,
\end{align*}
where classically $w>1$ for better faithfulness of $\boldsymbol{x}^{[0]}_\theta$ to $c$.
Often, e.\,g.\ in \citet{tashiro_csdi_2021}, it is sufficient to set $w=1$. This is equivalent to directly sampling from the conditioned model. Here, we call all of the above soft conditioning, because it does not enforce hard constraints on $\boldsymbol{x}^{[0]}_\theta$.

Inpainting is a method to impose hard constraints on $\boldsymbol{x}^{[0]}_\theta$. The name is derived from image diffusion: Some areas of an image are known in advance, and the general idea is to overwrite these areas with known values after each denoising step, effectively fixing them throughout the diffusion process. Meanwhile, the model adapts the generated, unknown parts of the image to match those areas \citep{lugmayr_repaint_2022, corneanu_latentpaint_2024}. The inpainting principle is also used in \citet{janner_planning_2022} to condition on start and potentially goal states when generating trajectories.

\paragraph{Notation}
The equations in this paper involve many different indices, for which we adapt the following notation: Denoising diffusion steps are denoted with a $t$, while time steps in an object trajectory are denoted with an $l$, and object indices with an $o$. We write indices of a larger tensor as subscripts (e.\,g.\ $\boldsymbol{z}_{o,l}$), steps in a computational process as superscript. Superscripts are either in square brackets when referring to diffusion steps (e.\,g.\ $\boldsymbol{x}^{[t]}$), or in parentheses when referring to intermediate layers in a neural network (e.\,g.\ $\boldsymbol{z}^{(1)}$).

\section{Object-centric Denoising Diffusion}

Our proposed object-centric denoising diffusion architecture models the time evolution of individual objects, as well as their interactions with each other, through a novel attention-convolution block (AC block). In the same way as \citet{janner_planning_2022}, we stack these blocks into residual blocks, which in turn are stacked to obtain a U-Net.
The whole architecture is shown in Figure \ref{fig:architecture} in the appendix.

To condition this model selectively on arbitrary states in different object trajectories, we propose a soft conditioning approach that combines masked-based interpolation with linear FiLM modulation \cite{perez_film_2018}. We additionally employ a hard conditioning approach based on trajectory shifting to anchor trajectories to pre-defined states.

\subsection{The Attention-Convolution Block}
The AC block, visualized in Figure \ref{fig:acb}, is comprised of a feed forward layer (MLP), a multihead attention layer (MHA) and a temporal convolution layer (Conv1D). This architecture entails translational equivariance over time, as well as permutation equivariance over objects, similar to real-world physics. As a consequence of this design, it allows training and inference with arbitrary numbers of objects and arbitrary trajectory lengths within the same model.

On a per-object, per-timestep level, the inputs to the AC block are $\boldsymbol{z}_{o,l}^{(0)} \in \mathbb{R}^{d_\mathrm{in}}$. These are vectors
that represent the state of object $o \in \{1, \shortdots, O\}$ at time $l \in \{1, \shortdots, L\}$.

The \textbf{MLP} layer serves two purposes: Projecting input features up to size $d_\mathrm{out}$ of the block's output, and producing features suitable for calculating interactions and temporal evolution in the following layers. It learns weights $\boldsymbol{W}_\mathrm{MLP} \in \mathbb{R}^{d_\mathrm{out} \times d_\mathrm{in}}$ and bias $\boldsymbol{b}_\mathrm{MLP} \in \mathbb{R}^{d_\mathrm{out}}$ to compute

\begin{align*}
\boldsymbol{z}_{o,l}^{(1)} = \text{MLP}(\boldsymbol{z}_{o,l}^{(0)}) = \text{Mish}\bigl(\text{GN}\bigl(\boldsymbol{W}_\mathrm{MLP} \boldsymbol{z}_{o,l}^{(0)} + \boldsymbol{b}_\mathrm{MLP}\bigr)\bigr) \, .
\end{align*}

For brevity, any Mish$(\cdot)$ activation function and GN$(\cdot)$ group norm operation are considered part of the overall layer. Grouping in GN$(\cdot)$ only concerns the $d_\mathrm{out}$ features of $\boldsymbol{z}_{o,l}$ and is therefore point-wise with regards to $o$ and $l$.

The \textbf{MHA} layer computes the interactions of object $\boldsymbol{z}_{o,l}^{(1)}$ with all objects $\bigl\{\boldsymbol{z}^{(1)}_{j,l}\bigr\}_{j=1}^{O}$ at a given time step $l$. It uses regular multihead self-attention \citep{vaswani_attention_2017} without positional encoding and is therefore permutation equivariant over objects. The MHA layer learns query, key and value matrices $\boldsymbol{W}_Q^h, \boldsymbol{W}_K^h, \boldsymbol{W}_V^h \in \mathbb{R}^{d_\mathrm{out}/H \times d_\mathrm{out}}$ for each head $h \in \{1, \shortdots, H\}$. It computes $\boldsymbol{z}_{o,l}^{(2)}$ as follows:

\[
\boldsymbol{q}^h_{o,l} = \boldsymbol{W}_Q^h\,\boldsymbol{z}^{(1)}_{o,l},\quad
\boldsymbol{k}^h_{j,l} = \boldsymbol{W}_K^h\,\boldsymbol{z}^{(1)}_{j,l},\quad
\boldsymbol{v}^h_{j,l} = \boldsymbol{W}_V^h\,\boldsymbol{z}^{(1)}_{j,l}
\]

\[
\alpha^h_{o,j,l} = \frac{\exp\bigl((\boldsymbol{q}^h_{o,l})^\top \boldsymbol{k}^h_{j,l}/\sqrt{d_\mathrm{out}/H}\bigr)}
                       {\sum_{j'=1}^{O} \exp\bigl((\boldsymbol{q}^h_{o,l})^\top \boldsymbol{k}^h_{j',l}/\sqrt{d_\mathrm{out}/H}\bigr)}
\]

\[
\boldsymbol{u}^h_{o,l} = \sum_{j=1}^O \alpha^h_{o,j,l}\,\boldsymbol{v}^h_{j,l}.
\]

\[
\boldsymbol{z}_{o,l}^{(2)} = \mathrm{MHA}(\boldsymbol{z}_{:,l}^{(1)}) = \text{GN}\bigl(\mathrm{Concat}\bigl(\boldsymbol{u}^1_{o,l},\shortdots,\boldsymbol{u}^H_{o,l}\bigr)\bigr)
\]

The colon denotes the inclusion of the whole range (here of objects). There is no activation function due to the subsequent skip connection.

Object interactions in physics are commonly modeled with graph neural networks \citep{melnik_benchmarks_2023}. We found that using graph attention GATv2 \citep{brody_how_2021} instead of MHA is considerably slower while generating trajectories of similar quality. GATv2 can handle sparse interaction graphs, while our MHA layer implicitly uses a fully-connected interaction graph at each timestep. However, such a fully connected graph is required for denoising in any case, even if actual object interactions are local and sparse, because in a noisy trajectory it is not yet clear which objects will interact with each other at what time step.

A \textbf{skip} connection allows bypassing the MHA layer. It concatenates $z_{o,l}^{(2)}$ with $z_{o,l}^{(1)}$ and projects the result back down to $d_\mathrm{out}$. Using this instead of an additive residual connection allows the model to learn how to combine both inputs, and in particular to discard interaction information from the MHA layer if it is deemed irrelevant. The skip connection learns $\boldsymbol{W}_\mathrm{skip} \in \mathbb{R}^{d_\mathrm{out} \times 2 d_\mathrm{out}}$ and $\boldsymbol{b}_\mathrm{skip} \in \mathbb{R}^{d_\mathrm{out}}$, and computes

\begin{align*}
    \boldsymbol{z}_{o,l}^{(3)} &= \mathrm{skip}\bigl(\boldsymbol{z}_{o,l}^{(2)}, \boldsymbol{z}_{o,l}^{(1)}\bigr) \\
    &= \text{Mish}\bigl(\text{GN}\bigl(\boldsymbol{W}_\mathrm{skip} \mathrm{Concat}\bigl(\boldsymbol{z}_{o,l}^{(2)}, \boldsymbol{z}_{o,l}^{(1)}\bigr) + \boldsymbol{b}_\mathrm{skip}\bigr)\bigr) \, .
\end{align*}

The \textbf{Conv1D} layer, lastly, computes the temporal evolution of each object $\boldsymbol{z}_{o,:}^{(3)}$ across all time steps $l$. It learns kernel slices $\boldsymbol{K}_s \in \mathbb{R}^{d_\mathrm{out} \times d_\mathrm{out}}$ as well as a bias $\boldsymbol{b}_\mathrm{Conv} \in \mathbb{R}^{d_\mathrm{out}}$, and calculates the output $\boldsymbol{z}_{o,l}^{(4)} \in \mathbb{R}^{d_\mathrm{out}}$ of the AC block as
\begin{align*}
    \boldsymbol{z}_{o,l}^{(4)} &= \mathrm{Conv1d}\bigl( \boldsymbol{z}_{o,:}^{(3)} \bigr)\\
    &= \text{Mish}\left(\text{GN}\left( \boldsymbol{b}_\mathrm{Conv} + \sum_{s=-k}^k \boldsymbol{K}_s \boldsymbol{z}_{o,l+s}^{(3)}\right)\right) \, .
\end{align*}
Conv1D uses zero-padding to preserve trajectory length, which is omitted in the equation for brevity. The convolution layer is translation equivariant over time.

While it is possible to use self-attention instead of convolutions to calculate temporal evolution, this would require some relative (rather than absolute) positional encoding of time steps to maintain true translational equivariance. Additionally, 
convolutions tend to smoothen object features across their trajectory, which is consistent with physical reality. Self-attention does not inherently do this.

\subsection{Conditioning}
To maximize flexibility of our approach, we seek to allow imposing conditions on arbitrary objects at arbitrary time steps. Conditions are provided as pre-defined state vectors $\boldsymbol{x}_{o,l} \in \mathbb{R}^d$ for the overall trajectory $\boldsymbol{X} \in \mathbb{R}^{O \times L \times d}$. Such conditions are naturally sparse, i.\,e.\ most objects do not have conditions at most time steps. 

\paragraph{Soft conditioning}
We integrate conditions into the U-Net through linear modulation, similar to FiLM \cite{perez_film_2018}. To handle the sparsity, we extend FiLM with an interpolation based on a mask. The mask encodes the presence of a condition for a given object at a given time. In the presence of a condition, a linear modulation of intermediate object representation $\boldsymbol{z}_{o,l}$ takes place, but in its absence $\boldsymbol{z}_{o,l}$ remains unchanged. This makes it possible to apply linear modulation only to specific objects and time steps.

Inside our U-Net, every two AC blocks are stacked into a residual block. For details, see \citet{janner_planning_2022}. We denote intermediate tensors between residual blocks as $\boldsymbol{Z} \in \mathbb{R}^{O \times L \times \tilde{d}}$. Each object has $\tilde{d}$ intermediate features. A second U-Net of identical architecture but independent weights predicts intermediate condition tensors. Its intermediate feature dimension however is $2\tilde{d}$, so that tensors can be split into factors $\boldsymbol{C}_m \in \mathbb{R}^{O \times L \times \tilde{d}}$ and biases $\boldsymbol{C}_b \in \mathbb{R}^{O \times L \times \tilde{d}}$. Likewise, a third independent U-Net with feature dimension 1 computes an intermediate mask tensor $\boldsymbol{M} \in [0, 1]^{O \times L \times 1}$. $\boldsymbol{M}$ is continuous and clipped to $[0,1]$ after each layer. Such an architecture with a continuous rather than discrete mask allows conditions to partially affect other objects or time steps in $\boldsymbol{Z}$.

Conditions are applied to $\boldsymbol{Z}$ between all residual blocks, and update it through
\begin{align*}
    \boldsymbol{Z}' = \bigl(\boldsymbol{M} \odot \boldsymbol{C}_m + (\boldsymbol{1} - \boldsymbol{M}) \odot \boldsymbol{1} \bigr) \odot \boldsymbol{Z} + \boldsymbol{M} \odot \boldsymbol{C}_b \, ,
\end{align*}
where $\odot$ denotes elementwise multiplication and $\boldsymbol{1}$ and $\boldsymbol{M}$ are expanded to match the dimensions of the other tensors.

We empirically find that it is sufficient to apply this conditioning only to the first two residual blocks of the downsampling part of the U-Net, so the networks that compute $\boldsymbol{C}_m$, $\boldsymbol{C}_b$ and $\boldsymbol{M}$ are truncated accordingly. We sample directly from the conditioned model, which is equivalent to $w=1$ in classifier-free guidance.

\paragraph{Hard conditioning}
The soft conditioning approach allows targeted injection of conditions into the U-Net, and is required for an informed denoising process. However, it does not strictly guarantee that generated trajectories adhere to these conditions.

To anchor trajectories to the provided conditions, we shift trajectories of individual objects to meet conditions. For objects with one condition, we shift all predicted time steps $\hat{\boldsymbol{x}}^{[0]}_{o,:}$ by the difference $\hat{\boldsymbol{x}}^{[0]}_{o,l} - \boldsymbol{x}^{c}_{o,l}$ between predicted vector $\hat{\boldsymbol{x}}^{[0]}_{o,l}$ and condition vector $\boldsymbol{x}^{c}_{o,l}$, at each step of the denoising process. For multiple conditions, the shift is linearly interpolated between them so that all are met. Time steps in between conditions then smoothly transition from one to the next. Objects without conditions are not shifted at all.

The anchoring operation is differentiable and applied after predicting tensor $\hat{\boldsymbol{X}}^{[0]}$. It requires a modified training loss

\[
\mathcal{L} = \mathrm{MSE}\left(\hat{\boldsymbol{X}}^{[0]}, \hat{\boldsymbol{X}}^{[0]}_\mathrm{shifted}\right) + \mathrm{MSE}\left(\hat{\boldsymbol{X}}^{[0]}_\mathrm{shifted}, \boldsymbol{X}^{[0]}\right) \, .
\]

The penalty term $\mathrm{MSE}\left(\hat{\boldsymbol{X}}^{[0]}, \hat{\boldsymbol{X}}^{[0]}_\mathrm{shifted}\right)$ computes the mean squared error between raw prediction and shifted prediction, incentivizing the model to predict correctly shifted trajectories. The regular diffusion loss term $\mathrm{MSE}\left(\hat{\boldsymbol{X}}^{[0]}_\mathrm{shifted}, \boldsymbol{X}^{[0]}\right)$ then calculates how well the shifted trajectories match the ground truth. Resulting trajectories are always guaranteed to obey all conditions.

Without any interactions, shifting an object trajectory across its coordinate system preserves its physical plausibility. With interactions this is not the case, but since the operation is performed after every denoising step, the model can iteratively improve the predicted interactions of shifted trajectories. While the deployed linear interpolation at first appears to be a crude method of transitioning between conditions, its behavior is easier to anticipate for the model than more involved interpolation methods. The model will eventually learn to make such shifts unnecessary. It is also computationally cheaper.

\section{Experiments}

We evaluate our model on the PHYRE benchmark scenarios. In the following, we first describe our setup and specify experimental details. Then we provide examples that showcase the use of multiple conditions per object, perform a comparison of model variants, and examine the effect of varying numbers of objects and trajectory lengths in the same model.

\subsection{Setup} We restrict the denoising to movable objects (red, blue, green, grey), and even in these only to the features that change (location and rotation). Fixed objects (black, purple), as well as immutable features (color and size) are not denoised. Still, the model has access to all objects and all features through its input $\boldsymbol{X}^{[t]} \in \mathbb{R}^{O \times L \times d}$, in which only changeable features of movable objects contain noise. Those features are then denoised in $\boldsymbol{X}^{[t-1]}$, while all others are copied over from $\boldsymbol{X}^{[t]}$. Consequently, only changeable features of movable objects in the initial $\boldsymbol{X}^{[T]}$ are sampled from Gaussian noise, while the remainder has to be pre-specified. Each individual denoising process thus requires prior specification of the number of objects and their colors and sizes.

To improve model efficiency, we replace one-hot encoded color in $\boldsymbol{X}^{[T]}$ by a binary value that encodes whether an object is movable. This reduces data dimensionality and removes information from model input that is physically irrelevant. After the denoising process, color is restored in $\boldsymbol{X}^{[0]}$.

There are two further issues with the PHYRE tasks that we solve by augmenting the training data: Firstly, the available PHYRE tasks cover only few possible initial locations for most objects. To improve this coverage, we augment the data by shifting each scene in the training data by a two-dimensional random location offset, sampled from $[-1, 1]^2$. Secondly, PHYRE objects bounce at the borders of their original location range $[0, 1]^2$, even though there are no floor or walls. This is physically not plausible and becomes a problem when adding an offset to the whole scene. We mitigate this by adding fixed bars around a scene, effectively boxing it in. Now, objects interact with other fixed objects instead of invisible walls when reaching scene boundaries. Together, these augmentations introduce an approximate learned location equivariance to the model, because interactions in the training data become largely independent of location. We empirically find that this is necessary for any generalization beyond the limited training trajectories.

Overall, there are 25 templates in PHYRE-B (numbered 0 to 24 and illustrated in Figure 7 of \citet{bakhtin_phyre_2019}), each with 100 tasks. The available actions (i.\,e.\ choosing location and size of the red ball to get trajectories from the simulator) are continuous, so there are infinitely many. For both training and evaluation, we sample random but valid actions. In the evaluations below, we distinguish train and test data based on whether a task was used in training. However, since the location and size of the red ball are always random, even an evaluation on a train task contains different object trajectories and collision events.

The evaluations report the root mean squared error (RMSE) between the trajectories generated by the simulator and those generated by the model.
During training, we randomly provide 0, 1 or 2 conditions at random time steps for each object. For evaluations that report RMSE values, modeled object trajectories are always conditioned on the initial states of the simulated trajectory; due to friction the dynamics are not strictly reversible, or in other words a condition at time $l$ does not unambiguously imply the exact dynamics prior to $l$. RMSE is only evaluated on changeable features (location, rotation) of movable objects, which are all normalized to $[0,1]$ in the training data.

\subsection{Multiple conditions}

\begin{figure}[ht]
\centering
\includegraphics[width=\columnwidth]{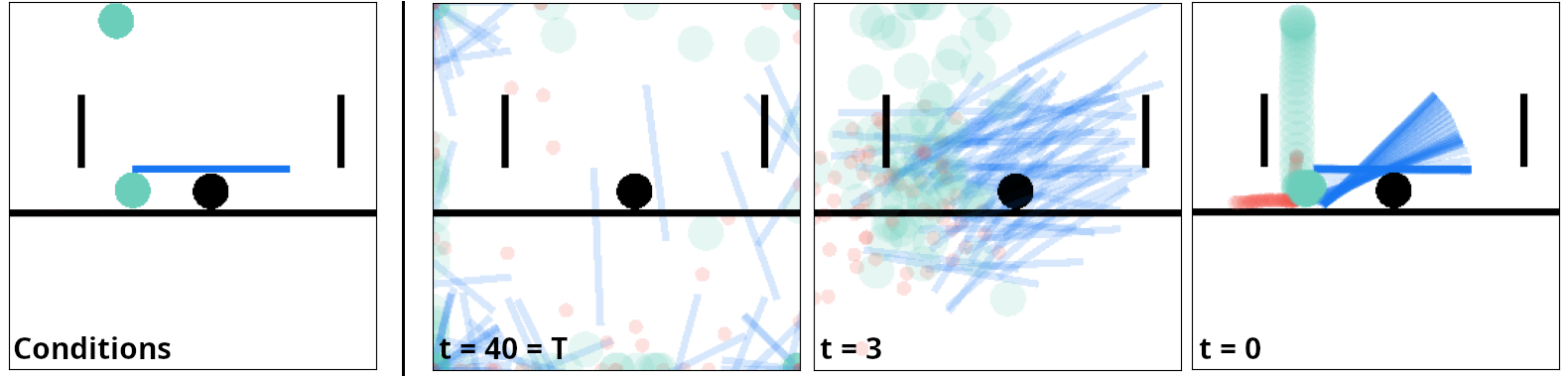}

\caption{\textit{Left:} Provided conditions fix initial states of green ball and blue bar, and the final state of the green ball. The red ball has no conditions. \textit{Right:} The denoising process.}
\label{fig:denoised_traj2}
\end{figure}

Figures \ref{fig:denoised_traj} and \ref{fig:denoised_traj2} show examples where conditions at multiple time steps are provided for the green and blue objects, while there are no conditions for the red ball. These conditions are motivated by the PHYRE goal of making green and blue objects touch. For visualization, object positions of partly denoised trajectories are clipped at the image boundaries. The model used for Figure \ref{fig:denoised_traj} is trained on all tasks of template 0, and the model used for Figure \ref{fig:denoised_traj2} on all tasks of template 4. The conditions are taken from those tasks, too.

These examples of generated trajectories show how our model is able to place the red ball in a way that the conditions are fulfilled, which is the main advantage of using a denoising diffusion model over autoregressive models in dynamics modeling. We empirically find, however, that the ability to generate plausible trajectories diminishes quickly for trajectories that are too far from the training data; e.\,g.\ if the red ball has a significantly different size than those seen during training on this task.

\subsection{Model Comparison and Ablation}

Comparing our model to related denoising diffusion approaches amounts to an ablation study. Approaches for trajectory generation in reinforcement learning and robotics encode the entire scene as a single feature vector and lack an object‑centric perspective \citep{janner_planning_2022,carvalho_motion_2024,ajay_is_2022}. To construct a comparable non‑object‑centric ablation (which we refer to as \textit{CNN only}), we modify our U‑Net by removing the MHA layer (and its skip connection) from the AC blocks and replace the per‑object Conv1D operations with a single convolution over the full state vector $\boldsymbol{z}_{:,l}\in\mathbb R^{O \cdot d_{\mathrm{out}}}$. In multi-agent tracking, \citet{ye_diffusion_2024} present an architecture very similar to ours, but since agents are only encoded through location, they lack the feature-processing MLP layer. We remove this layer from our model to create a \textit{No MLP} ablation.

Lastly, no other approaches use our anchoring for hard conditioning. The related approaches above rely on inpainting for conditioning. Empirically, we find that inpainting only fixes those time steps that were pre-specified, without affecting the rest of the trajectories. It is therefore virtually similar to removing the hard conditioning, which we also test. We are not aware of comparably flexible soft conditioning approaches that we could test against.

\begin{figure}[ht]
\centering
\includegraphics[width=0.9\columnwidth]{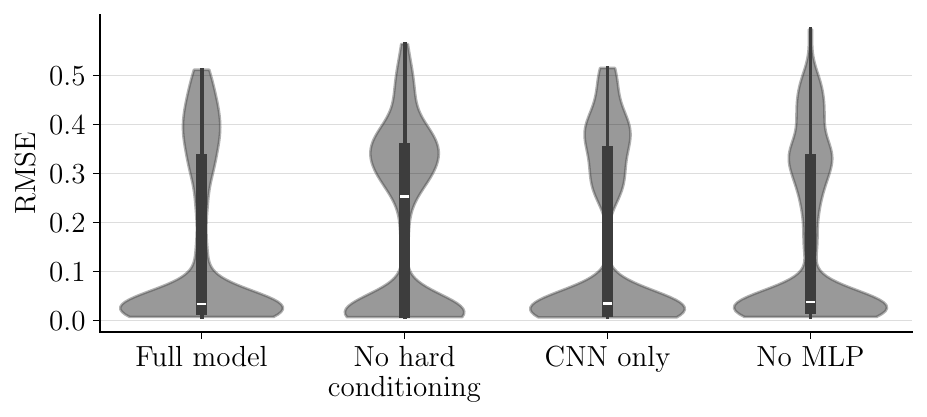}

\caption{Performance of model variants for trajectories of 3 interacting balls (PHYRE template 0), with $L=64$.}
\label{fig:comparison}
\end{figure}

Each model is trained on template 0, which contains 3 moving, interacting balls, and the resulting RMSE distributions are presented in Figure \ref{fig:comparison}. It becomes apparent that removing the hard conditioning significantly reduces performance. Removing either the MHA or the MLP leads to comparable median RMSEs (0.034, 0.035 and 0.038 for full model, \textit{CNN only} model and \textit{No MLP} model, respectively). 

However, the mean RMSE for the full model (0.153) is significantly better than for the \textit{CNN only} model (0.166) and \textit{No MLP} model (0.165). The mean RMSE without hard conditioning is 0.201.
The \textit{CNN only} model additionally has theoretical disadvantages: It can only be trained on a fixed number of objects, is not permutation equivariant and its number of weights scales linearly with the number of objects, while that number stays constant for the other models.

All violin plots have two clear modes, one below and one above RMSE $=0.15$. Case-by-case evaluation shows this is related to the number of object interactions, which increase the chance of mispredictions -- resulting either in physically implausible trajectories, or more often in slight deviations over multiple time steps, which add up in the RMSE. The waist of the distribution is however not a clear separation between scenarios with and without interactions; some of each appear below and above it.

\subsection{Number of Objects}

\begin{figure*}[ht]
\centering
\includegraphics[width=\textwidth]{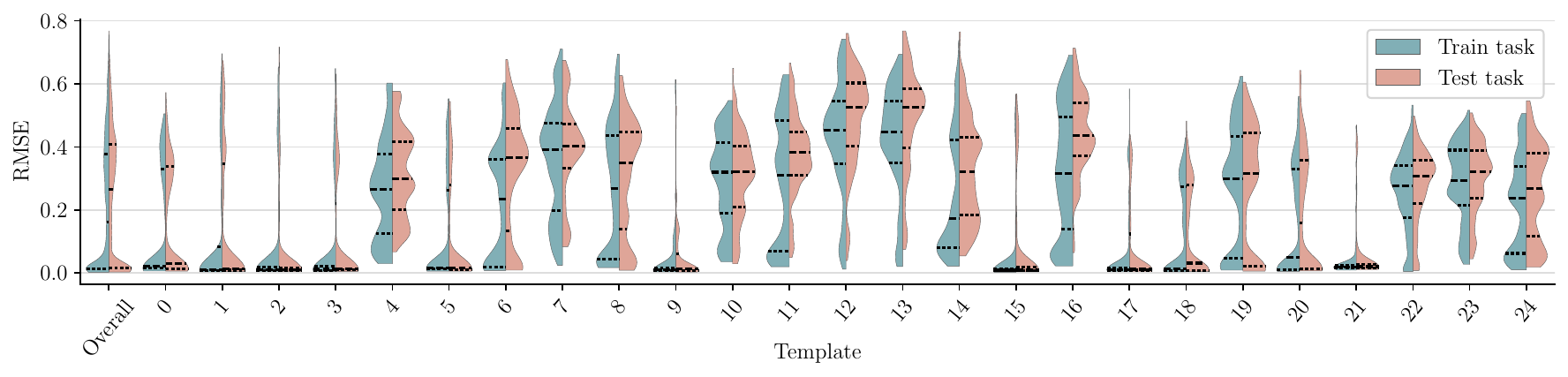}

\caption{Results when training a model on all 25 templates simultaneously. Performance on train tasks is blue, performance on test tasks red. The leftmost violin summarizes overall statistics. The black dotted lines indicate distribution quartiles.}
\label{fig:within_template}
\end{figure*}

To investigate the implications of using our model with varying numbers of objects, we train it on all 25 templates simultaneously. The PHYRE-B setup contains pre-defined folds, splits of train and test tasks for each template. Training on those train tasks takes about twice as long to converge as training on only template 0, using the same hardware.

In Figure \ref{fig:within_template} we evaluate this model on train tasks and on test tasks. It becomes apparent that some tasks achieve significantly better RMSEs than others. An inspection of the templates in \citet{bakhtin_phyre_2019} again shows that this correlates with the number of object interactions that take place.

Since there is a relatively small number of object types, and since object interactions are very local in time and space, the model should ideally be able to quickly learn about object interactions and generalize them across templates. The differences in performance between templates suggests that this generalization capability is limited. A likely cause is that the denoising process has to consider all possible object interactions across the entire scene and temporal evolution. We suspect that the model considers all object positions relative to each other during denoising, preventing an efficient generalization of interactions. In PHYRE in particular, the available relative object locations across tasks are relatively sparse, so the model can attribute individual interactions to individual task configurations rather than local object behavior. Presumably, a better coverage of relative object positions and interaction patterns in the training data could at least partially mitigate this.

\subsection{Trajectory Length}

\begin{figure}[ht]
\centering
\includegraphics[width=\columnwidth]{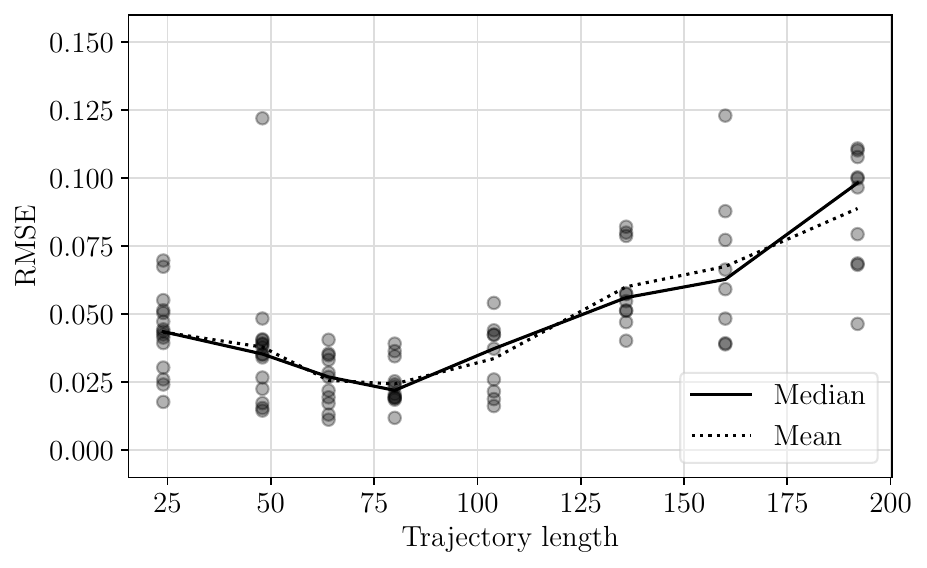}

\caption{Quality of generated trajectories for different trajectory lengths $L$. Based on the full model trained on template 0 with 3 interacting balls. Training data has $L=64$. Only RMSE values $<0.15$ are considered.}
\label{fig:length}
\end{figure}

Finally, we explore model behavior when generating trajectories of various lengths. Our models are trained on trajectories of length $L=64$, and we evaluate our full model, trained on template 0, for different values of $L$. The resulting RMSEs are reported in Figure \ref{fig:length}. Since we want to analyze the best-case scenario, we consider only RMSE values $<0.15$, but note that the resulting pattern is similar for the upper part of RMSE distributions.

The results show that deviating from training length gradually reduces performance. Qualitatively, this often means that objects still move smoothly, but not correctly. They might move through each other, or suddenly start falling upwards. We believe that this is caused by the temporal abstraction happening in the U-Net \citep{janner_planning_2022}. It introduces boundary effects into the Conv1D layer that the model learns to handle specifically for the trajectory lengths it sees during training.

\section{Related Work}

\citet{daniel_ddlp_2023} use denoising diffusion models to predict the movement of objects. Their model is designed in an object-centric way, but it is unconditioned, and instead of a U-Net, it relies on an autoregressive transformer architecture. If the model was conditioned, this autoregressive architecture would likely stand in the way of imposing conditions on later time steps of generated trajectories.

There are various works on trajectory generation with denoising diffusion models. In reinforcement learning, \citet{janner_planning_2022, ajay_is_2022} and others (see \citet{zhu_diffusion_2024} for an overview) use denoising diffusion models for generating plausible agent trajectories. These models use inpainting for conditioning, which does not work well in our scenario. Our U-Net architecture is based on that of \citet{janner_planning_2022}, although their version (as well as others) does not disentangle environment states into individual objects.

In agent tracking, models generate trajectories of multiple moving agents. \citet{jiang_motiondiffuser_2023} already propose a permutation equivariant architecture. It is based on self attention but compares whole agent trajectories, not feature vectors at distinct time steps. The work of \citet{ye_diffusion_2024} is perhaps closest to ours. Their U-Net architecture is very similar to ours, but lacks the MLP layer, because the agents are specified only by location, rather than by arbitrary feature vectors. Additionally, their conditioning is based on information from before the generated time horizon, and thus cannot be conditioned on time steps within generated trajectories.

In general time series prediction, \citet{tashiro_csdi_2021} employ an architecture with a similar modularity to ours: One transformer module computes interactions between time series, and another computes the evolution of each time series.

In a different direction, \citet{xu_interdiff_2023} use denoising diffusion to generate human-object interaction scenarios, and obtain plausible interactions by using a physical correction step within the denoising process. This is possible because each denoising step predicts an $\boldsymbol{x}^{[0]}$, to which corrections can be applied. Beyond our trajectory anchoring, we have attempted something similar in two ways: We have introduced a term to the loss that penalizes predicted object overlaps, and, going a step further, we have implemented a differentiable correction module that resolves overlaps after each denoising step. However both models have failed to converge to a reasonable prediction quality, and we leave this very relevant avenue to future work.

\section{Conclusion}

In this work, we present a novel conditioned, object-centric denoising diffusion model for physical reasoning. It generates trajectories of multiple, interacting objects, each of which can be conditioned on arbitrary time steps within its trajectory. To realize this conditioning, we introduce a masked interpolation approach to infuse conditions into the denoising network, as well as a trajectory shifting approach to enforce conditions in the generated trajectories. The whole architecture is translationally equivariant over time as well as permutation equivariant over objects.

We demonstrate how our model can solve physical reasoning tasks where some object trajectories are conditioned on multiple time steps, while other objects can be placed freely. This is relevant, for instance, to generate object trajectories that connect initial conditions to desired goal states. Previous physical reasoning approaches are often autoregressive and cannot be conditioned on goal states, while previous denoising diffusion models are not designed for object-based physical reasoning.

A current limitation is our inability to test how the limited coverage of relative object positions in training data impacts generalization. This analysis would require a physical reasoning benchmark with sufficiently diverse object configurations and access to object-level features. To the best of our knowledge, such a benchmark does not exist yet. We leave this for future work on more suitable benchmarks.

\section*{Acknowledgements}
This research was supported by the research training group ``Dataninja'' (Trustworthy AI for Seamless Problem Solving: Next Generation Intelligence Joins Robust Data Analysis) funded by the German federal state of North Rhine-Westphalia.

\bibliography{paper}

\appendix

\section{Appendix}

\subsection{U-Net architecture}
\label{appendix:architecture}
\begin{figure*}[ht]
\centering
\includegraphics[width=\textwidth]{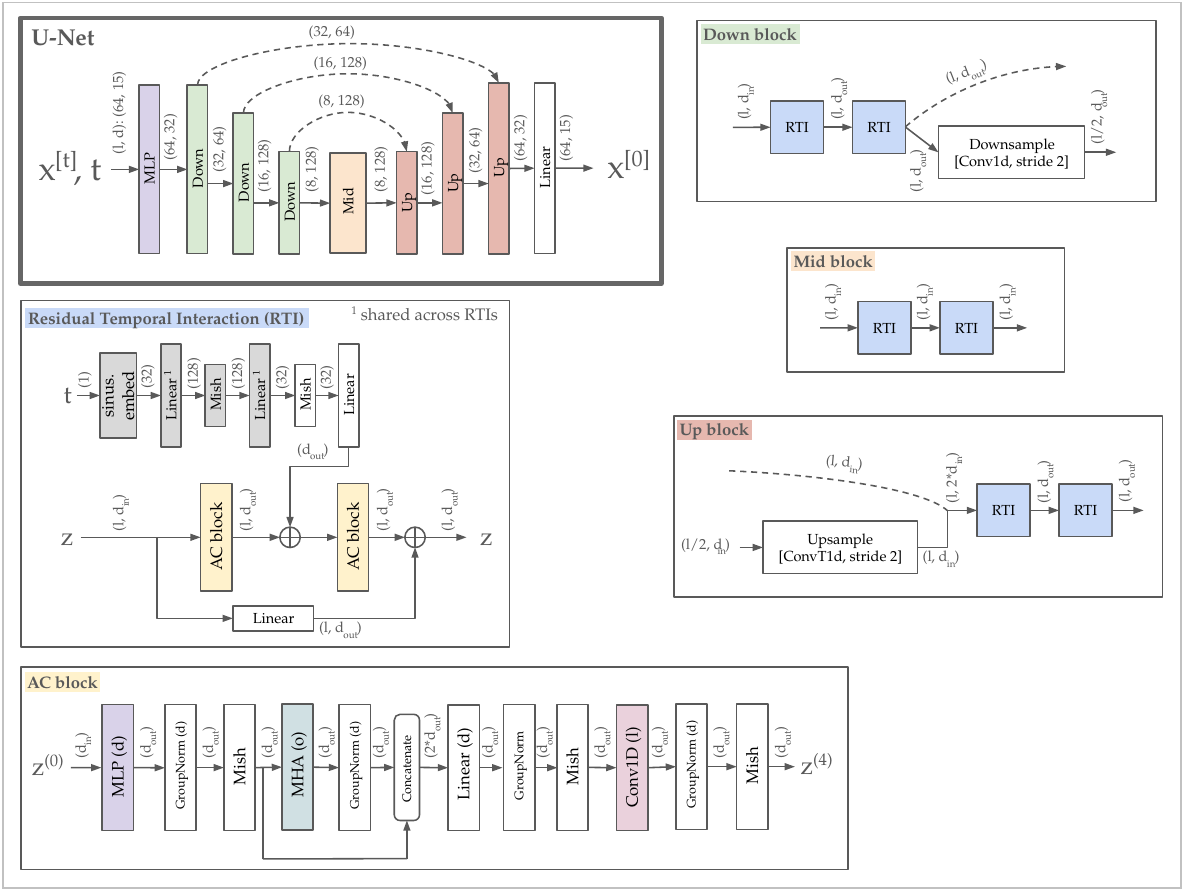}

\caption{Full architecture of our proposed denoising diffusion model, without conditioning modules. The values for parameters $o, l, d$ are the ones that are used for experiments reported in this paper. In the AC block, the letters $o, l, d$ indicated with each element indicate the dimension of $z$ across which this element operates. The colors only serve the purpose of visually distinguishing components.}
\label{fig:architecture}
\end{figure*}

The full architecture of the U-Net we use is presented in Figure \ref{fig:architecture}. The values provided for $l$, $o$ and $d$ are those we used for the experiments in this paper. The presented model does not include the conditioning models that calculate $C_m$, $C_b$ and $M$. Those are two truncated copies of the presented architecture with different values for $d$, and their outputs are fed into the main architecture after each residual temporal interaction (RTI) block. They are trained jointly with the main network through these connections.

\end{document}